\documentclass[11pt]{article}
\usepackage[T1]{fontenc}
\usepackage[utf8]{inputenc}
\usepackage{amsmath,amssymb}
\usepackage{graphicx}
\usepackage[margin=1in]{geometry}
\usepackage{booktabs}
\usepackage{threeparttable}
\usepackage{pifont}
\usepackage{subfig}
\usepackage{xcolor}
\usepackage{hyperref}
\usepackage{natbib}
\setcitestyle{authoryear,round}

\renewcommand{\checkmark}{\ding{51}}
\newcommand{\xmark}{\ding{55}}

\title{Structuring Quantitative Image Analysis with Object Prominence}

\author{
Christian Arnold\\
\small School of Government, University of Birmingham, Birmingham, BT15 2TT, United Kingdom
\and
Andreas K\"upfer\thanks{Corresponding author: \texttt{andreas.kuepfer@tu-darmstadt.de}}\\
\small Institute for Political Science, Technical University Darmstadt, Darmstadt, 64283, Germany
}
\date{}

\begin{document}

\maketitle

\begin{abstract}
When photographers or media professionals compose an image, they make deliberate choices about what to foreground and what to background, shaping how viewers interpret visual content. Yet most quantitative approaches to image analysis overlook this structure and treat detected objects as equally important. We introduce a framework for measuring object prominence\textemdash the relative salience of objects in an image\textemdash as a means to make computational image analysis attentive to the compositional emphasis a curator has built into an image. Drawing on research in cognitive psychology and computer vision, we outline three approaches for estimating object prominence: size and centeredness, inferred depth, and saliency maps. Validating that curator-composed prominence measurably shifts human visual attention in a pre-registered eye-tracking study, we illustrate this framework's benefits in two further applications. First, we demonstrate how weighting features in line with their prominence can enhance the unsupervised ideological scaling of U.S. newspaper images. Second, we examine gendered visual prominence in U.S. presidential campaign ads from 2016 and 2020, showing that Republican candidates depict women less prominently than their Democratic counterparts. Our framework lets researchers analyze image data at scale while remaining attentive to its communicative structure and intent.
\end{abstract}


\section{Introduction} 

Images have become an increasingly valuable form of data in political research, yet much of the computational infrastructure for their analysis remains underdeveloped compared to text \citep[e.g.,][]{joo2022image, torres2022learning, WWilliams2020images}. This gap is particularly apparent when it comes to interpreting the relative importance of different objects within an image. Although recent methods in political science, building on advances in object detection and local keypoint description from the computer vision literature \citep{arandjelovic2012three, wu2019detectron2}, successfully identify people, objects, and other relevant elements in images \citep{Scholz_2025, Torres2024Vbow}, they typically assume that all detected elements are equally important. This is a limiting assumption. Human viewers rarely interpret visual content in such a flat way. Instead, they attend more to some objects than to others: eye-movement studies show that fixations on objects in real scenes are jointly shaped by low-level visual emphasis\textemdash size, spatial position, contrast\textemdash and by an object's semantic importance to the scene \citep{PARKHURST2002107, tHart2013fixations}. Computational models of salient object detection formalize this intuition, estimating salience directly from an image \citep{Borji_2019, Zhou_2021}. This oversight in current image analysis frameworks is not just technical. It has direct implications for how we model and interpret images as forms of political communication. In images that have been consciously composed\textemdash for instance, by media outlets or political campaigns\textemdash choices about what to foreground and what to relegate to the background are often deliberate and meaningful \citep{kress2020reading, Yang24}. By failing to account for these choices, conventional approaches risk overlooking important aspects of how visual content is used to persuade, inform, or mislead.

We argue that quantitative image analysis stands to benefit from attending to how images are composed, rather than only focusing on what they contain. Viewers do not scan every object in an image with equal interest: both bottom-up visual salience and top-down goals jointly guide where the eyes land \citep{rayner200935th}, and objects that are visually distinctive or spatially prominent are more likely to draw an early fixation \citep{PARKHURST2002107}. Curators exploit this and highlight some objects more than others. Building on established methods from computer vision \citep{Borji_2019, Zhou_2021} and theoretical insights from cognitive psychology, our proposed framework separates the task of identifying objects in an image from the task of estimating the compositional emphasis placed on them. We validate our approach against human attention in a pre-registered eye tracking experiment. 

The framework opens up a variety of analytical possibilities. In some cases, object prominence can be used to refine existing workflows for image analysis. For example, in studies that infer ideological positions from image corpora, it may be helpful to weigh the presence of objects by how likely they are to be noticed. A campaign advertisement featuring a national flag prominently may carry a different ideological signal than one in which the flag is relegated to the background \citep{Joo_2014_CVPR, Torres2024Vbow}. In other cases, object prominence is the primary focus of analysis itself. Consider research questions about how visibility is enhanced or diminished for certain groups in political media. Studies in text-based political communication have already identified systematic patterns in group representation, such as the underrepresentation of women in news media and parliamentary debates \citep{ozer2023women, shor2015media, pasgender2020}. Our framework allows researchers to extend these inquiries to visual content.

The remainder of the paper proceeds in five parts. Section 2 introduces the concept of object prominence and situates our contribution within the broader literature on image-as-data approaches \citep{loken2021using, Scholz_2025, Torres2024Vbow}, building on recent advances in computer vision while offering new tools for social scientific inquiry. Section 3 outlines three canonical approaches to measuring object salience, each differing in complexity, interpretability, and implementation cost. Section 4 tests whether these measures track human attention in a pre-registered eye-tracking study. Section 5 demonstrates the empirical value of the framework through two applications: the first uses object prominence to improve unsupervised ideological scaling of news outlets \citep{Scholz_2025, Slapin2008Wordfish}, while the second treats prominence as the outcome of interest, analyzing gendered visual prominence in U.S. presidential campaign advertisements. Section 6 concludes.

\section{Object Prominence: Curation as a Structuring Principle for Image Analysis}

The images-as-data literature comprises three strands: defining semantic units in images for further analysis \citep{joo2022image, loken2021using, Scholz_2025, schwemmer2023automated, Torres2024Vbow, williams2024, WWilliams2020images}; modeling latent, theory-based image characteristics instead, such as visual aesthetics \citep{peng2022athec}; and predicting outcomes of interest directly from images as a whole, where neural networks learn representations without annotating predefined objects and the prediction model itself is a black box \citep{joo2022image, torres2022learning, WWilliams2020images}.\footnote{Vision Transformers would belong to this category, too.}

We contribute to the first strand of this literature: our framework extends the work of those interested in analyzing semantically meaningful objects in images. In a broad array of applications, social scientists are keen to analyze such data at scale, and to do so in a way that respects how it was produced. Over the last few decades, political science has successfully embraced close consideration of the data-generating process when modeling structured data \citep{Aldrich08, brauninger2020theory, king1998unifying}. Some have transferred these ideas to working with unstructured data, most prominently text \citep{egami2022make, Slapin2008Wordfish} and audio data \citep{knox_lucas_2021}. Following these footsteps, we take the first steps in extending this work to visual data.

Quantitative analyses of images often start by detecting and delineating objects within each frame, whether faces and demographic attributes \citep{joo2022image}, annotated objects in the study of political violence \citep{loken2021using}, or object and facial recognition more generally as a critical first step \citep{schwemmer2023automated, Scholz_2025, WWilliams2020images, williams2024}. \citet{Torres2024Vbow} takes a different, inductive angle: rather than identifying pre-defined objects, it clusters similar local keypoints, which the analyst then interprets and assigns meaning.

``Just'' knowing an image's semantic units already enables meaningful insights, much as the bag-of-words assumption has for text \citep{laver2003extracting, roberts2014structural, Slapin2008Wordfish}. Several projects use image objects this way to infer political ideology or affiliation, whether from an image's objects directly \citep{xi2020understanding}, from facial landmarks \citep{joo2015automated}, or from multimodal social media content \citep{wang2017polarized}. 

Our starting point is that politically relevant images are rarely accidental, but instead curated. A photographer chooses a vantage point and a moment; a picture editor selects one image from many and decides how it is cropped and placed; a campaign designs an advertisement frame by frame. News production is a process of selecting and arranging content \citep{shoemakervos2009gatekeeping}, and visual material is no exception: picture editors act as visual gatekeepers who decide which images reach the public and how they are presented \citep{schwalbe2015gatecheckers}. Scholars of political communication have long shown that such choices matter, demonstrating how the visual packaging of candidates and issues shapes their portrayal \citep{grabebucy2009image, messaris2001role, kress2020reading}. In the coverage of contentious politics, for instance, the selection and composition of protest imagery can lend or withhold legitimacy from collective actors \citep{corrigallbrown2012picturing}. For this reason, we move beyond the assumption that every detected object carries equal weight. Describing image content is a prerequisite for interpreting it \citep{loken2021using}: Annotation can be paired with questions about who produces and interprets an image \citep{williams2024} and graph models can relate individuals, objects, and their environment \citep{Yang24}. We build on this work by making the relative weight of objects explicit and measurable.

Curators express their intent through \emph{visual emphasis}: the compositional choices that make some objects dominant and others peripheral. A curator can enlarge an object, move it toward the center of the frame, bring it into the foreground, or set it against a contrasting background. These are the levers of emphasis, and they are encoded in the image itself. We therefore suggest that analyzing an image involves two steps. First, it is necessary to detect the areas of interest in an image and turn them into variables for analysis. These areas may be predefined objects recovered by a supervised model, or regions that an unsupervised approach groups together and to which the analyst then assigns meaning \citep{Scholz_2025, Torres2024Vbow}. Second, we ask how strongly the composition emphasizes each area. The object prominence is the product of these two components,

\begin{equation}
    \text{Object Prominence} = \text{Area of Interest} \times \text{Visual Emphasis}.
\end{equation}

The two components enter multiplicatively by design: an area that is entirely absent receives no prominence regardless of how the rest of the image is composed, so that emphasis modulates only the areas that are actually present. Among present areas, prominence increases with emphasis: an area relegated to the background carries little, while one that is foregrounded carries much. In corpus applications, these per-area prominences replace the raw presence or counts that conventional approaches feed into an object-document matrix, so that each area contributes in proportion to the emphasis the image places on it rather than being treated as equally important \citep{Slapin2008Wordfish, Torres2024Vbow}.

Three distinctions clarify what visual emphasis is and is not. First, visual emphasis is not the same as framing. Framing is the broader act of selecting aspects of reality and making them salient in order to promote a particular interpretation \citep{entman1993framing}, and visual framing operates at several levels, from the denotative content of an image to its connotative and ideological meaning \citep{rodriguezdimitrova2011levels}. Visual emphasis is narrower and more concrete: it is the compositional layer that governs which objects a composition foregrounds. It is one input to visual framing rather than a synonym for it, and we use the term in this restricted sense throughout.

Second, visual emphasis is not the attention mechanism of a neural network. Vision transformers learn internal attention weights end-to-end, optimized to improve prediction on a specific task \citep{dosovitskiy2020image}. Those weights highlight whatever features help the classifier, which need not correspond to anything a human arranged or would notice. Our measure differs in both target and interpretation: it recovers the compositional choices encoded by the image's creator, and it is defined independently of any downstream prediction task.

Third, and most importantly, we make a claim about the image, not about the viewer. We measure the emphasis a curator has built into a composition. We do not claim that every viewer attends to the same regions, nor that our measure reproduces any individual's perception. Viewers differ, and their attention is shaped by prior beliefs, interests, and context. That heterogeneity is precisely why the alignment between curated emphasis and viewer attention is an empirical question rather than an assumption. Eye tracking offers a direct way to study viewer attention in political science \citep{Jenke_Sullivan_2025}, and we treat this alignment as a matter for validation rather than something to be assumed.

Our framework is well suited to images produced with communicative intent, such as press photographs, campaign advertisements, and political imagery on social media. It is less relevant where images arise without deliberate composition, as with a surveillance camera. Where a curator is at work, the arrangement of the scene is a purposeful act: the sender composes the image so that its message is likely to be received as intended.\footnote{Images are polysemous and always open to multiple readings. Qualitative analysis remains important for recovering context-dependent meaning \citep{barthes1967elements, chandler2022semiotics, kress2020reading}.} This intentionality is documented directly in the two substantive domains we validate the framework on below: climate-change photojournalism is shaped by deliberate, shifting visual-framing choices on the part of editors and photographers \citep{oneill2019climatechange}, and political campaigns visually differentiate candidates by gender, including through the facial prominence given to women relative to men \citep{jungblut2023visual, valmori2021facial}.

Making visual emphasis explicit yields analytical leverage in two ways. In some studies, prominence serves as a preprocessing step, refining measures that would otherwise treat every object alike, as in the ideological scaling of news imagery \citep{Scholz_2025, Slapin2008Wordfish, Torres2024Vbow}. In others, prominence is the object of study itself, letting researchers ask how visibility is granted to or withheld from particular groups, extending existing text-based work on the underrepresentation of women \citep{ozer2023women, shor2015media, pasgender2020} to the visual domain \citep{gasparyan2026pixels, karekurve2026missing}. Our two applications below illustrate these two uses of a measurement tool: It allows researchers to improve ideological scaling, e.g., a corpus of newspaper articles. Moreover, object prominence can itself be the theoretical construct of interest. Beyond these two cases, the same logic extends to other settings where a curator's compositional choices carry political meaning. For example, researchers could compare not just whether demonstrators, police, or counter-protesters appear in protest coverage but how prominently each is composed, extending existing work on how coverage can lend or withhold legitimacy to collective actors \citep{corrigallbrown2012picturing}. In a similar vein, it would allow scholars to study how state outlets cover their respective head of state relative to their entourage as a scalable indicator of personalization in leadership coverage, extending work on the communicative intent behind political imagery \citep{Joo_2014_CVPR}.

\section{Operationalizing and Measuring Object Prominence}

Measuring object prominence has two parts. The first is to define the set of areas of interest in an image, i.e., to determine which pixels belong to a semantically meaningful entity \citep{schwemmer2023automated}: analysts can train a supervised classifier \citep{torres2022learning}, deploy or fine-tune a pre-trained model \citep{peng2023automated, Scholz_2025}, or cluster local key points without predefined categories \citep{Torres2024Vbow}. 

Our concern in this section is the second part: estimating the visual emphasis the composition places on each area of interest. How strongly does the composition make each one stand out? Computer vision offers many tools for this, ordered from the most transparent to the most involved: \emph{size and centeredness}, \emph{foreground and background}, and \emph{salient object detection}. They differ in technical complexity, interpretability, computational cost and in how closely they are built to track human attention \citep{Barcelo2020complexity, Linardatos2020}. Since research projects have different requirements, we offer concrete guidance on which to choose.

\begin{figure}[h!]
    \centering

    \begin{minipage}[t]{0.24\textwidth}
        \includegraphics[width=\linewidth]{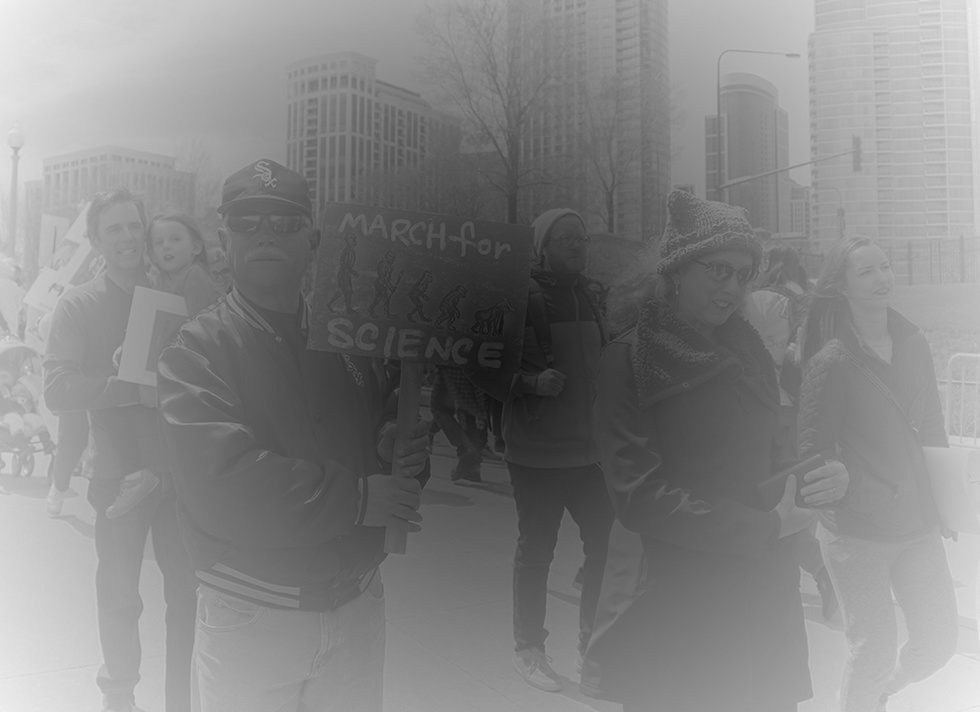}
        \centering (a) Centrality Map
    \end{minipage}
    \hfill
    \begin{minipage}[t]{0.24\textwidth}
        \includegraphics[width=\linewidth]{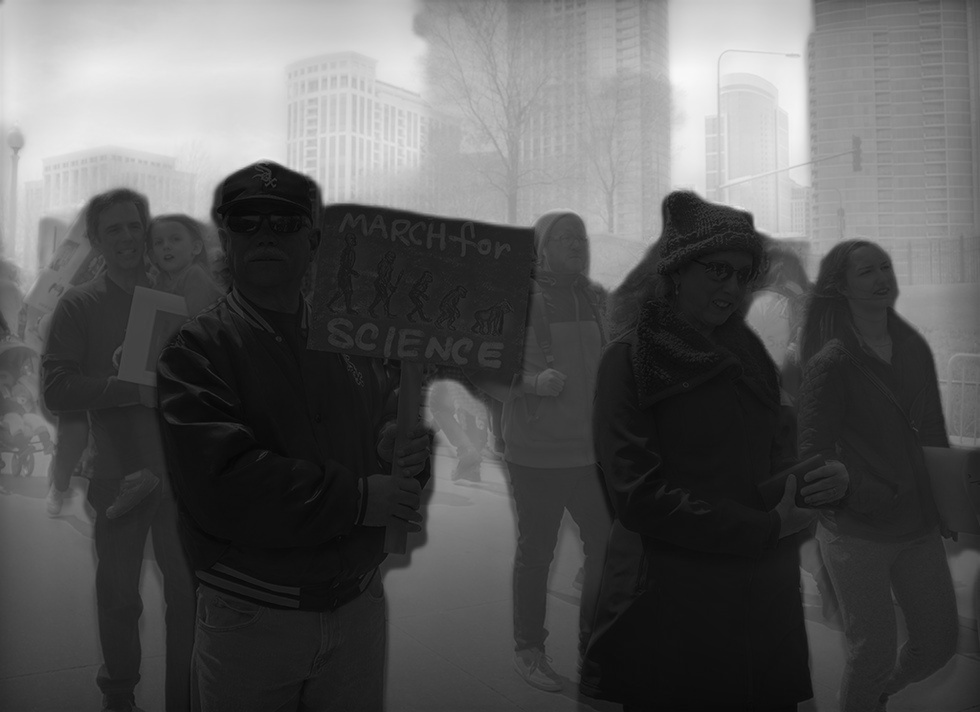}
        \centering (b) Depth Map with ZoeDepth \citep{bhat2023zoedepth}
    \end{minipage}
    \hfill
    \begin{minipage}[t]{0.24\textwidth}
        \includegraphics[width=\linewidth]{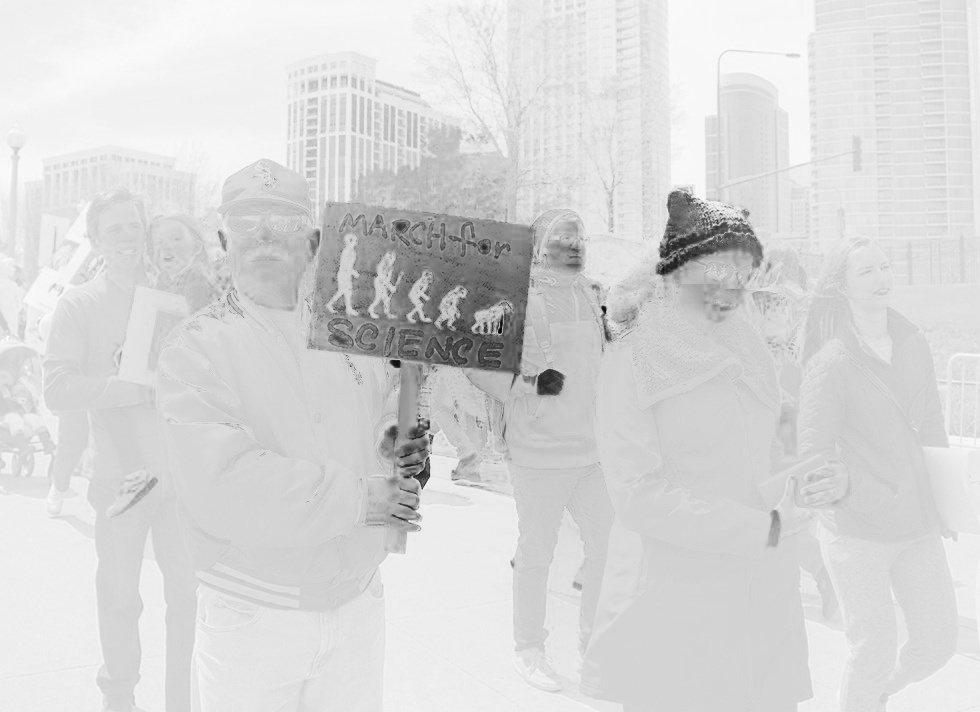}
        \centering (c) Salient Object Detection with MBD \citep{Zhang2015minimum}
    \end{minipage}
    \hfill
    \begin{minipage}[t]{0.24\textwidth}
        \includegraphics[width=\linewidth]{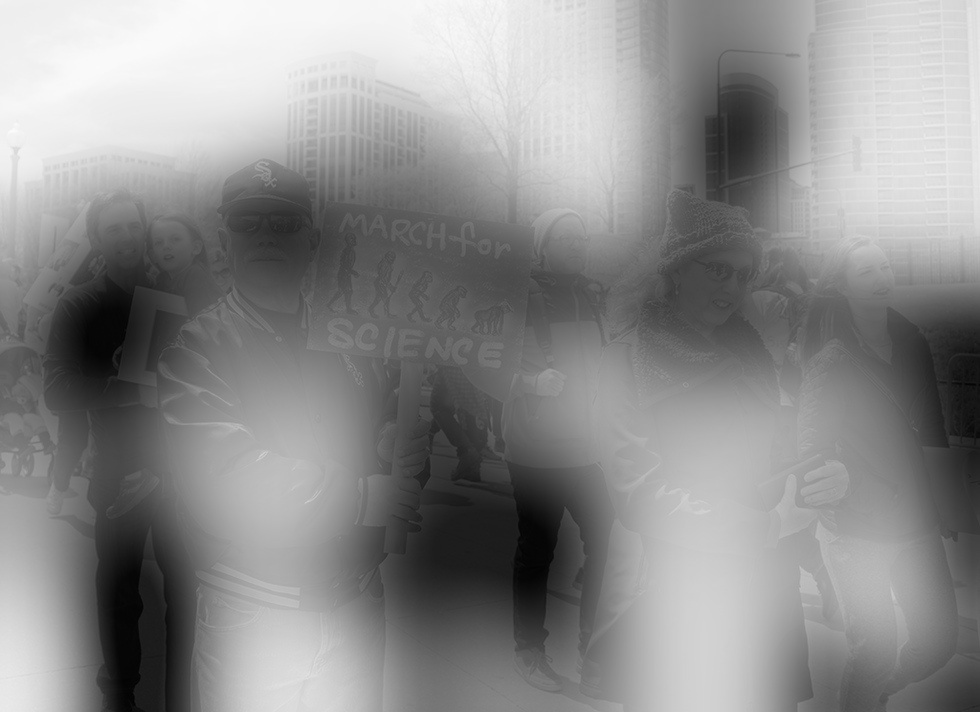}
        \centering (d) Salience Map with \citet{itti1998saliency}
    \end{minipage}

    \caption{Comparison of different object prominence approaches. Darker areas indicate higher relative visual emphasis.}
    \label{fig:examples}
\end{figure}

The first approach is the computationally cheapest, most transparent, and easiest to interpret, which makes it attractive for large corpora. It rests on two cues a curator manipulates directly: how large an object is and how central it sits. Size is the ratio of the object's area to the area of the image. Centeredness is the Euclidean distance between the object's centroid and the geometric center of the frame. Creators enlarge and center the objects they wish to emphasize, and these choices leave measurable traces in the image \citep{kress2020reading, grabebucy2009image}. Objects of interest tend to lie near the center in standard image collections \citep[][123]{Borji_2019}, and larger objects tend to be judged more important \citep[e.g.,][]{berg2012importance, spain2011measuring}. Both features can be read off segmentation masks or bounding boxes produced by standard models \citep{zhao2015, goekhan2015size, wu2018size}. Figure~\ref{fig:examples}a shows a centrality map, where darker central areas indicate higher emphasis.

The second approach adds a third spatial dimension. A depth map records the distance from the camera to each pixel, letting the analyst distinguish objects a curator has placed in the foreground from those relegated to the background \citep{Lang2012depth}. Recovering depth from a single image once relied on geometric and photometric cues such as perspective, shadow, and focus \citep{Saxena2007depth}. Modern models learn it from large training sets and apply readily to a new corpus \citep{ming2021depthsurvey}.\footnote{Stereo-based methods using two parallel cameras also exist \citep{laga2022depthsurvey, Saxena2007depth}.} Depth is more involved than size or centeredness, but its models, although freely available, can be less transparent \citep[e.g.,][]{bhat2023zoedepth, Yang_2024_CVPR}. Figure~\ref{fig:examples}b shows a depth map.

The third approach is the most involved and, from a conceptual standpoint, most directly targets what we mean by emphasis. In vision science, \emph{salience} is the property that makes a region stand out from its surroundings and draw attention independently of the viewer's momentary goal. It is standardly represented as a \emph{saliency map} that topographically encodes the conspicuity of every location in a scene \citep{ittikoch2001, koch1985shifts}. Methods of salient object detection (SOD) estimate such a map: a continuous, per-region score of how strongly each part of the image stands out \citep{Borji_2019, Zhou_2021}. Every such algorithm reads only the image, but the algorithms differ in what they are built to reproduce. One strand consists of explicit models of human visual attention, engineered so that the saliency map approximates where people actually look. The classic example is \citet{itti1998saliency}, which combines low-level contrast, color, and intensity into a map of the areas most likely to attract the eye \citep{itti2000saliency, koch1985shifts}. More recent deep-learning models predict human fixations far more accurately and come with usable open implementations \citep{linardos2021deepgaze, lou2022transalnet}. A second strand is engineered instead to segment the compositionally most salient object as a clean mask, tuned against object-segmentation benchmarks rather than against gaze. A representative and efficient example is Minimum Barrier Detection \citep[MBD;][]{Zhang2015minimum}, a fully unsupervised technique that treats regions connected to the image border as background and isolates the foreground object through color and texture contrast \citep{jiang2013backgroundnesscue}.\footnote{Other widely used, well-documented salient object detection models with open implementations include U$^2$-Net \citep{qin2020u2net} and BASNet \citep{qin2019basnet}.} Figure~\ref{fig:examples}c shows an MBD map, and Figure~\ref{fig:examples}d shows the corresponding map from \citet{itti1998saliency}'s gaze-oriented model, the same algorithm we validate against human attention in Section~4.

We offer explicit practical guidance for choosing among these approaches, since there is no generically ``best'' solution: the right choice depends on the research question. Size and centeredness are the cheapest and most transparent, and are the default choice when computational cost or interpretability is paramount. Depth adds the foreground--background dimension at moderate additional cost, and can offer more diagnostic value than size or centeredness alone on images with complex spatial composition. Salient object detection is the most involved, and here the choice of algorithm matters: when alignment with human attention specifically is the goal, a gaze-mimicking model such as \citet{itti1998saliency} is the appropriate choice. Algorithms tuned for compositional segmentation instead, such as MBD, are appropriate when the goal is isolating the foreground segments rather than modeling gaze (Appendix~\ref{app:validation}).

\section{Validating Object Prominence Against Human Attention}
\label{sec:validation}

Even if our framework primarily focuses on how actors compose images, these compositions should have behavioral downstream consequences. Eye movements are a direct behavioral record of where a viewer looks, and they therefore serve as a human ground truth against which a computational measure of prominence can be judged. Since some algorithms for salient object detection are explicitly designed to capture human attention, that claim can be tested directly. If a curator composes an image so that certain objects stand out, those objects should draw the attention of the people who view it. We examine this with a pre-registered eye-tracking study.\footnote{See \url{https://osf.io/bxpgd/overview}.}

For the first experiment, we selected seven press photographs of social protest\footnote{six from the 2019--2020 \emph{Estallido Social} in Chile and one from a 2020 protest in the United States} and altered each one: in three we added a prominent object in the foreground, and in four we removed one and refilled the space. Each image therefore exists in two versions that differ only in a single prominent object.\footnote{Adding or removing a whole object shifts both whether an object is present and how strongly the composition emphasizes it, so Experiment~1 validates object prominence as a whole rather than isolating the emphasis component on its own. It does not, however, collapse into the near-tautology that an added object attracts the eye, since the test concerns the graded distribution of salience across the grid rather than the mere presence of an object. Isolating the emphasis component directly is a natural next step for future work.} Fifty-four participants each viewed every image.\footnote{Participants were staff and students at the University of Birmingham. Three sessions were excluded for poor recording quality, leaving 54 participants in Experiment~1. Each image was shown for ten seconds.}

To compare human attention with the algorithm, we imposed a uniform 8x8 grid on every image and treated each cell as a unit of observation. Within each cell we computed the algorithm's mean saliency score \citep{itti1998saliency}, and for each image pair the change in that score between the two versions, $\Delta\text{CVScore}$. A cell containing the altered object shows a large $\Delta\text{CVScore}$, a background cell almost none. We then measured in three ways how much attention each cell drew: through the number of fixations, the total dwell time, and the number of returns.\footnote{A fourth pre-registered measure, first fixation, did not show reliable alignment (see Appendix~\ref{app:validation}). The first fixation is a single, very noisy event, and would require a far larger sample than the count measures. We read the null as a limit of statistical power, not as evidence against the measure.} 

Because condition varies between participants, H1 is tested as an interaction: participants in the treatment condition should attend more than those in the control condition precisely in the cells where the algorithm predicts a larger rise in prominence. We model each measure with a negative binomial regression, including fixed effects for participant, image, and cell (coefficient table in Appendix~\ref{app:validation}). Figure~\ref{fig:val_h1} presents the result as simulated first differences \citep{kingtomz2000making}. For all three measures the difference rises with $\Delta\text{CVScore}$ at a very similar slope and intercept. It crosses the line at around $\Delta\text{CVScore} = 0$ and turns clearly positive where the algorithm predicts the largest increase: where the algorithm sees an object become more prominent, viewers in the treatment condition fixate it more often, dwell on it longer, and return to it more (interaction coefficients of $0.17$, $0.19$, and $0.18$, all $p<0.001$).\footnote{The pre-registered coarser 4x4 grid yields the same pattern, Appendix Table~\ref{tab:val_h1_itti}, but with slightly increased standard errors. The interaction remains significant on both grids.} We find that curated prominence measurably steers attention. This alignment does depend on how visual emphasis is measured: we also implement the MBD algorithm as robustness check \citep{Zhang2015minimum}. While the interaction effect is positive, too, it is not significant at our sample size.\footnote{See Appendix~\ref{app:validation} for further details.}

\begin{figure}[htbp]
    \centering
    \includegraphics[width=\linewidth]{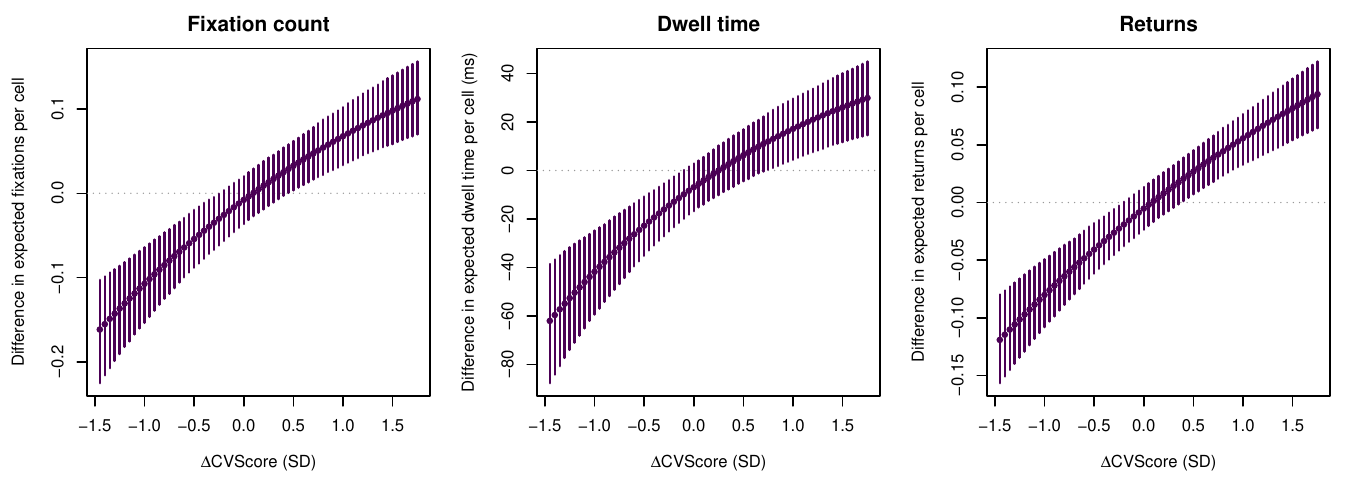}
    \caption{Experiment 1. Simulated first difference in expected human attention (treatment minus control) as a function of algorithm-predicted prominence change, $\Delta\text{CVScore}$ (Itti model, $8\times8$ grid). Points are expected first differences with 95\% intervals. Attention rises where the algorithm predicts greater prominence.}
    \label{fig:val_h1}
\end{figure}

Taken together, our experimental validation shows that algorithms designed to mimic visual emphasis are actually doing so in practice. When a curator alters the prominence of an object, human attention shifts accordingly. This is direct behavioral evidence that composed visual emphasis is not merely a property analyst impute, but one that shapes how people actually look at it. The validation does not show that the measure reproduces any individual's full perception, nor that every operationalization performs equally well. But attention-grounded measures of emphasis align with human gaze, and that alignment does not appear to hinge on the viewer's own politics.\footnote{A second experiment found no evidence that a viewer's own political ideology reshapes this attention (Appendix~\ref{app:validation}).}

\section{Analytical Leverage from Applying Object Prominence}
Including salience maps in computer-based image analysis introduces significant analytical leverage, particularly in research projects that work with human-curated content, such as news photographs or political advertisements. Object prominence opens the door to different applications, and we focus on two examples here. First, we show how to improve the idealpoint estimation of newspapers. Second, we focus on a case where object prominence itself is at the center of the study, examining the prominence of women in campaign ads for the 2016 and 2020 U.S. presidential electoral races.

\subsection{Prominence as a Preprocessing Method: Improving Idealpoint Estimation}
In this first study, we utilize object salience as a means to improve the pre-processing of image corpora \citep{denny2018text}.

Transforming image data to a machine-readable and intuitive format is challenging. While a text-based corpus comes with words as natural semantic units, images are a collection of pixels, where this relationship still has to be established. Recent work has shown how to use supervised and unsupervised methods to convert image corpora into data representations similar to term-document matrices, where a spreadsheet would count similar semantic units in its rows for each image in columns \citep{Scholz_2025, Torres2024Vbow}. Image corpora that are converted into such a data representation can be studied in a manner similar to text. 

We demonstrate how object prominence can enhance analyses based on such data representations. Our use case comes from ideological scaling. Ideological positions are an essential concept that allows understanding the preferences of political actors. Retrieving this information from text sources has marked a significant step forward in analyzing politics across space and time \citep{laver2003extracting, Lowe2008Wordscores, Slapin2008Wordfish}. Following studies of media's ideological positioning \citep{gentzkow2010drives, turkel2021method}, we study newspaper slant. We rely on the data from \citet{Thomas2019Predicting}, who crawled a comprehensive database of political newspaper articles from sources marked as left/right, or extreme left/right. Their data covers almost 500 media outlets, focused primarily on the U.S.. Among the topics available in their data, we focus on climate change: it constitutes a clear, well-documented ideological cleavage in U.S. politics, while also being a highly salient and widely covered issue beyond the U.S. context. From their data, we select eight news outlets (Wonkette, Advocate, Esquire, CNN, Breitbart, Newsmax, Fox News, and RedState) with an even split between left and right-leaning sources. Our sample spans a range of ideological intensity on each side rather than only contrasting left against right: CNN and Fox News are mainstream cable outlets, while Breitbart, RedState, and Wonkette are more overtly partisan blogs. Moreover, we introduce variation in format, combining national cable news with political blogs and niche lifestyle magazines (Esquire, Advocate), so that our test case is not limited to a single kind of media outlet. Out of a pool of 2,853 climate change articles available across these outlets, we draw a random stratified sample of 120 articles with text and a photo. We instruct two independent research assistants to annotate the articles based on their text and place them on an ideological scale ranging from -5 (far left) to 5 (far right). Our ground truth for each article is the average of the two annotators' scores.\footnote{We rely on text as our ground truth because it was already used to classify the corpus into issues \citep{Thomas2019Predicting} and is easy to interpret ideologically. Appendix~\ref{sec:codebook} gives the codebook, intercoder reliability, external validation, and a discussion of text/image divergence.} This data allows us to calculate the ground truth of each outlet's average ideological position on a left-right scale as a measure of media slant.

Our key innovation is to scale the newspapers with the \emph{Wordfish} model \citep{Slapin2008Wordfish} using an object-document matrix that is refined with object salience. As a first step, we identify the relevant semantic units in the model. We use two different approaches: The unsupervised detection of key points with the \textit{RootSIFTdescriptor} model \citep{arandjelovic2012three, Torres2024Vbow} and the supervised detection of objects with the \textit{Detectron2} model \citep{Scholz_2025, wu2019detectron2}. 

Instead of taking the raw count of these objects in each image, we introduce weights. When working with text, adding weights to the term-document matrix (e.g., with term frequency-inverse document frequency) has been shown to improve many downstream tasks \citep{sparck1972statistical}. We apply the same logic to images: each detected feature or object is weighted by its own salience score rather than counted uniformly.

We measure object salience with Minimum Barrier Detection (MBD) \citep{Zhang2015minimum}, and additionally report all results using the alternative gaze-validated algorithm from \citet{itti1998saliency} as a robustness check.

We construct these weights differently for the two approaches. In the unsupervised approach, weights enter during k-means clustering of extracted keypoints: each detected keypoint is assigned to one of 2,000 visual-word clusters that make up a shared, corpus-wide vocabulary, weighted by its own salience score. This means a keypoint on a prominent, foregrounded object contributes more to its cluster than one on a receding background, even if both appear equally often in the raw image. In the supervised approach, we build an analogous matrix from named object categories, weighting each identified object by its own average salience score. In both cases, these per-image values are summed within each outlet across the full pool of that outlet's climate-change images to form the outlet-by-object matrix Wordfish scales: $\log \lambda_{ij} = \alpha_i + \psi_j + \beta_j \theta_i$, where $\theta_i$ is outlet $i$'s position on the latent left-right dimension, $\psi_j$ captures how frequently object $j$ appears overall, and $\beta_j$ captures how strongly it discriminates between left- and right-leaning outlets\textemdash the object-level analogue of a word's discriminating power in text scaling. Appendix~\ref{sec:app1-hyperparams} walks through a fully worked example for both approaches. Overall, we can thus compare four different scenarios: the unsupervised \citep{Torres2024Vbow} and supervised \citep{Scholz_2025} approaches to detect points of interest, with and without the respective objects weighted according to their salience.

Table~\ref{tab:scaling_mae} reports results for all scenarios. We calculate the mean absolute error between the models' idealpoints and the human-annotated ground truth, taken as the average of two independent annotators (Appendix~\ref{sec:codebook}). The Wordfish model, based on plain unsupervised data representation, has an MAE of 1.252. Introducing weights reduces the MAE to 0.661 under MBD, and to 0.695 under the gaze-validated \citet{itti1998saliency} algorithm. When relying on the supervised approach to annotate the images, the Wordfish model based on plain data representation has an MAE of 0.560. Introducing weights increases the MAE to 1.054 under MBD, and to 0.833 under Itti\textemdash worse than the unweighted baseline under either algorithm.

Object prominence almost halves the MAE in unsupervised image analysis\textemdash a clear improvement\textemdash while in the supervised case it does not improve the results and in fact makes them worse. The supervised pattern may be rooted in a general property of supervised object detectors: region-proposal architectures filter candidate detections by confidence, and transformer-based architectures apply attention across image regions during classification; in either case, objects that are small, peripheral, or otherwise low-signal are already harder for the detector to confidently identify, and are more likely to be down-weighted or discarded before results are returned. An explicit salience weight added on top of what survives this internal filtering mostly reintroduces noise rather than signal, so our results argue against further weighting for prediction in this setting, but that narrows only prominence's role as a predictive booster. Section~5.2 shows it is equally useful as an outcome of interest in its own right, independent of prediction, which a presence-only measure could not have captured.

\begin{table}[ht]
\centering
\begin{tabular}{lcccc}
  \hline
Scenario & Object Prominence? & Salience Algorithm & MAE (Mean) & SD \\
  \hline
  Unsupervised  & \xmark & --- & 1.252 & 0.449 \\
  Unsupervised  & \checkmark & MBD \citep{Zhang2015minimum} & 0.661 & 0.040 \\
  Unsupervised  & \checkmark & Itti et al. (1998) & 0.695 & 0.022 \\
  Supervised  & \xmark & --- & 0.560 & 0.013 \\
  Supervised & \checkmark & MBD \citep{Zhang2015minimum} & 1.054 & 0.085 \\
  Supervised & \checkmark & Itti et al. (1998) & 0.833 & 0.050 \\
   \hline
\end{tabular}
\caption{Average Mean Absolute Error (MAE) from hand coding (average of two independent annotators) for scaling estimates per scenario. Average includes a set of different minimum and maximum term frequency parameters for Wordfish. The unweighted scenarios do not depend on a salience algorithm; the weighted scenarios are reported under both salience algorithms used elsewhere in the paper (Section~\ref{sec:validation}).}
\label{tab:scaling_mae}
\end{table}

\subsection{Prominence as a Substantial Method: Gender Representation in U.S. Presidential Campaign Videos}

In this second application, we treat object salience as the \emph{explanandum}, examining how visually prominent women are made in political advertisements during the 2016 and 2020 U.S. presidential campaigns. Voting decisions are shaped by group representation and identification with political actors \citep{cutler2002simplest, lewis2009american, Persson_2015, mondak1995votemedia, erksen1990voters}, and gender is a particularly influential dimension of this \citep{rosenthal1995role, Phillips_2018}: when women perceive themselves as marginalized in a party's communication strategy, this can weaken their identification with it. While much attention has been paid to the gender characteristics of political candidates, we suggest that gendered representation also extends to the broader visual and communicative context.

A growing body of research examines the visibility of women in political texts, including speeches, news coverage, and social media \citep[e.g.,][]{ozer2023women, shor2015media, pas2022media, pasgender2020}. These studies consistently find that women are underrepresented in textual political communication. Our approach extends this line of inquiry to the visual domain, and studies the question using a dataset of 1,934 video advertisements released by candidates during the 2016 and 2020 U.S. presidential campaigns \citep{fowler2020presidential, fowler_2023_2020}. We are particularly interested in whether and how prominently Democrats and Republicans feature women in their video ads.

To answer our research question, we follow our proposed object prominence framework and identify the women in videos, as well as their visual emphasis. We begin by extracting individual scenes from all video ads: each frame is converted to grayscale, and the mean absolute pixel-wise difference to the preceding frame is computed. Whenever this difference exceeds a threshold (30, on a 0--255 grayscale intensity scale), the frame is saved as a new, unique scene.\footnote{We rely on \url{https://github.com/montoulieu/frame-slice} and apply scene-change detection to 1,917 videos, yielding 40,370 scene frames in total. Not every ad catalogued by WMP had a locally available file, and some processed files (chiefly Spanish-language variants) are not catalogued by WMP as separate ads, so this count does not subtract exactly from the 1,934 catalogued ads. See Appendix~\ref{sec:app2-data} for the full accounting and more details.} Next, we detect the bounding box of faces from images and classify their gender using a pre-trained model \citep{serengil2024lightface}. Salience weighting does not require multiple competing objects to be informative. Even in a single-face frame, how a candidate is composed, shown as a large, close-up figure that dominates the shot, or as a small, distant figure in a wide shot, still carries a meaningful signal about visual emphasis. A campaign that chooses to depict a candidate one way rather than the other is itself a communicative decision worth measuring, regardless of whether other objects are present.\footnote{Beyond this conceptual point, single-face frames are also not the norm in our data, with a long tail of crowd and rally scenes containing dozens of faces (Appendix Figure~\ref{fig:faces_per_frame} reports the full distribution).} We then measure the salience of each face in each scene of a video in two ways. As a first measure, we compute the image depth map in each image \citep{bhat2023zoedepth}. With this, we can calculate the inverted frame-normalized depth of detected faces, where a lower value indicates that the face is allocated in the background of an image. As a second measure, we rely on the relative size of faces and calculate the relative area of a frame occupied by a particular face.

\begin{figure}[h]
    \centering
    \includegraphics[width=.8\linewidth]{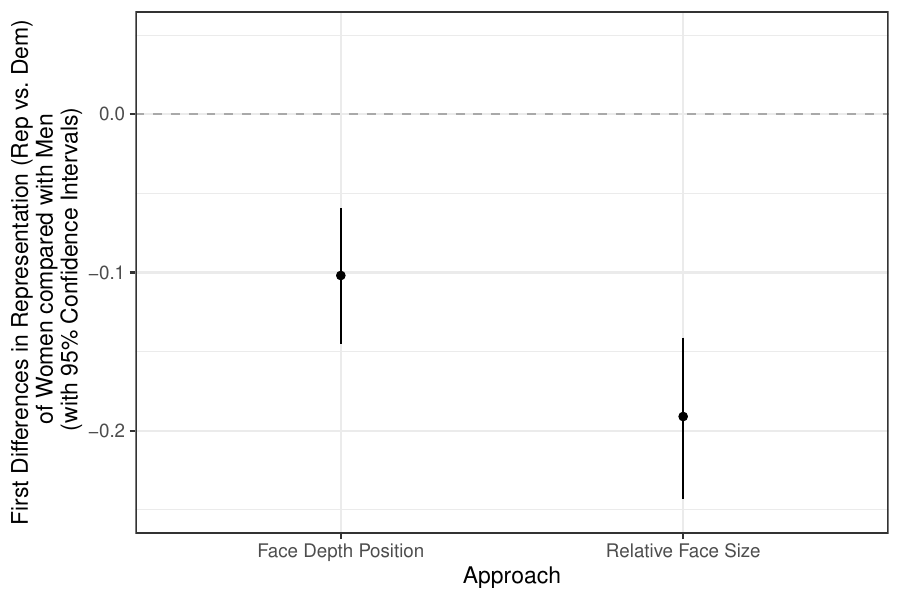}
    \caption{First differences in expected representation of women vs.\ men, by party. Face Depth Position uses normalized face depth as outcome. Relative Face Size uses relative face size, both \textit{z}-standardized with fixed effects for candidate, election year, and visibility (64,406 / 58,674 faces. see Appendix~\ref{sec:app2-data}).}
    \label{fig:results}
\end{figure}

To examine how political parties visually communicate gender, we estimate two regression models that capture different dimensions of object prominence. Both models include fixed effects for candidate ID, election year, and the candidate's visual presence in the video. Each model interacts the annotated gender of a detected face with the political party associated with the advertisement. The first model uses the normalized depth of a face within the frame as the outcome variable, capturing how close or far individuals appear in the visual field. The second model employs the relative size of a detected face, measured as a proportion of the overall frame. To facilitate direct comparison of effect sizes, both outcome variables are \textit{z}-standardized.

As shown in Figure~\ref{fig:results}, the estimated first differences reveal consistent patterns: Republican candidates, on average, depict women as less visually prominent in their ads compared to Democrats. For the model with the normalized depth, the effect is most likely at -0.10. For the second model, it is most likely at -0.19. In both models, the interaction effects are statistically significant at the 95\% confidence level.

What does this add beyond whether women appear in campaign ads at all? Both models include a fixed effect for a candidate's overall visual presence, so the gap reflects not screen time but how prominently women are composed once they appear\textemdash large and close, or small and distant. Existing text-based work on women's underrepresentation \citep{ozer2023women, shor2015media, pas2022media, pasgender2020} detects only whether women are mentioned or shown at all. It cannot detect this second, compositional form of marginalization, which a presence-only measure would miss entirely and which is exactly the sense in which this application validates the broader framework.

\section{Conclusion}

This paper introduces a framework for measuring object prominence in quantitative image analysis: the compositional emphasis a curator places on the objects present in an image, distinct from their mere presence. Drawing on computer vision and cognitive psychology, we operationalize this emphasis through size and centeredness, depth, and salient object detection, and we show in a pre-registered eye-tracking study that curator-composed emphasis measurably shifts human attention. By offering a modular, scalable method for weighting visual elements by their likely prominence, we aim to reduce measurement error and enhance the interpretability of image-as-data analyses.

We demonstrated this framework's utility through two applications. First, in scaling newspapers' ideological positions from their imagery, weighting points of interest with their visual emphasis nearly halves the error of an unsupervised pipeline, but does not improve\textemdash and in fact worsens\textemdash a supervised one, suggesting that a pretrained object detector's own confidence already implicitly reflects prominence. Second, in campaign advertisements from the 2016 and 2020 U.S.\ presidential elections, Republican ads compose women as less visually prominent than Democratic ads do, a gap that holds conditional on a candidate's overall visibility and so reflects marginalization in how prominently women are shown once included, not merely in whether they appear.

Beyond these two applications, the framework applies wherever a curator's compositional choices carry political meaning, whether as a preprocessing step or as an outcome of interest in its own right (Section~2). It is compatible with both unsupervised and supervised object detection pipelines, though our results show that the benefit of salience weighting differs across them and should be assessed case by case rather than assumed. As image-based data continue to grow in political science, we believe this framework offers researchers a principled guidance to move beyond treating images as flat collections of objects, toward studies that take seriously how those objects are composed.

\paragraph{Acknowledgments}

Jana Bernhard, Deren Onursal, Dylan Paltra, Zachary Steinert-Threlkeld, Michelle Torres, and Nils Weidmann provided tremendously helpful comments to a previous draft of the paper. The authors acknowledge support by the state of Baden-Württemberg, Germany, through bwHPC, and the Birmingham Transformative Humanities Lab. We thank Daniel Kuhlen, Yuhan Luo and Rupali Limachya for research assistance. Earlier versions of the draft were presented at the 2024 COMPTEXT meeting at the Vrije Universiteit (VU), Amsterdam, May 2-4, 2024, at EPSA's 14th annual conference in Cologne, July 4-6, 2024, at APSA's 120th Annual Conference, Philadelphia, Pennsylvania, September 5 to September 8, 2024, at the LOOPS Workshop: ''How Image-As-Data Approaches Can Help Analysing Protest and Its Organisation: Methods and Applications'', Free University Berlin, September 30 to October 1, 2024, at the Friday Seminar Series, Trinity College Dublin, March 14, 2025, and at the Annual Conference of the DVPW Section Methods of Political Science, Mannheim, March 27-28, 2025

\paragraph{Data Availability Statement}

Replication data and code for both applications and the validation study will be made available in a public repository upon acceptance, in accordance with the journal's data-access and research-transparency requirements.

\paragraph{AI Usage Statement}

This paper uses a variety of Artificial Intelligence (AI) tools, including \textit{Detectron2} (Application 1), \textit{ZoeDepth} (Application 2), and \textit{DeepFace} (Application 2). The authors used Claude (Anthropic) for copyediting (e.g., identifying typographical errors and suggesting wording improvements to reduce length).

\newpage
\appendix

\setcounter{table}{0}
\renewcommand{\thetable}{A\arabic{table}}
\setcounter{figure}{0}
\renewcommand{\thefigure}{A\arabic{figure}}

\newpage

\setcounter{page}{1}
\begin{center}
    \begin{LARGE}
    \textbf{Structuring Quantitative Image Analysis with Object Prominence\\}
    \end{LARGE}
    \vspace{.25cm}
    Christian Arnold (University of Birmingham)\\ Andreas Küpfer (Technical University of Darmstadt)
    \vspace{1cm}

 Appendix
\end{center}

\renewcommand{\thesection}{\Alph{section}}

\section*{Table of Contents}

\begin{itemize}
    \item[A.] Application 1: Ideological Annotation of Climate Change Texts (Codebook, External Validation, and Ground-Truth Discussion)
    \item[B.] Application 1: Matrix Construction, Hyperparameter Selection, and Robustness Checks
    \item[C.] Application 2: Data Processing and Regression Results
    \item[D.] Validation: First Fixation, Algorithm Sensitivity, and Deviations
\end{itemize}

\newpage

\section{Application 1: Ideological Annotation of Climate Change Texts (Codebook, External Validation, and Ground-Truth Discussion)}\label{sec:codebook}

Each text in the dataset pertains to the topic of \textit{climate change}. Your task is to assign an \textbf{ideology score} that reflects the text’s perceived position on the left–right political spectrum.

\subsection*{Scoring Scale}

Use an 11-point scale ranging from \textbf{–5 (far left)} to \textbf{+5 (far right)}:

\begin{itemize}
    \item \textbf{–5}: Strongly left-wing. Explicitly frames climate change as an urgent crisis requiring aggressive government intervention (e.g., a Green New Deal, binding emissions targets, fossil fuel bans); attributes blame to corporations, industry lobbying, or right-wing politicians; quotes primarily activists, environmental scientists, or progressive officials.
    \item \textbf{–2 to –4}: Left-leaning. Endorses climate action and regulation without the strongest rhetoric above, or is critical of industry/conservative policy but in a more measured tone, or relies mostly (not exclusively) on left-coded sources and framing.
    \item \textbf{–1}: Slight left lean. A mostly factual or descriptive piece with a mild tilt toward pro-climate-action framing, e.g., in word choice or which voices are quoted first or at greater length.
    \item \textbf{0}: Neutral, balanced, ambiguous, or ideologically indeterminate (see ``Handling Ambiguous, Mixed, and Insufficient-Information Cases'' below for how to distinguish these).
    \item \textbf{+1}: Slight right lean. Mirrors –1, but tilting toward skepticism of climate policy or sympathy for industry.
    \item \textbf{+2 to +4}: Right-leaning. Skeptical of climate regulation or the urgency of climate change, favors market solutions over government mandates, or relies mostly on industry/conservative sources and framing.
    \item \textbf{+5}: Strongly right-wing. Explicitly dismisses or minimizes climate change as exaggerated, a hoax, or not human-caused; frames climate policy as government overreach harmful to jobs or the economy; quotes primarily industry representatives or conservative politicians/commentators.
\end{itemize}

\subsection*{Indicators to Consider}

Base your score on the following, in order of priority, moving to the next only if the current one does not resolve the score: (1) \textbf{explicit policy positions} (e.g., support for or opposition to emissions targets, carbon pricing, fossil fuel restrictions); (2) \textbf{framing of climate change itself} (settled and urgent vs.\ contested or exaggerated); (3) \textbf{attribution of blame} (corporations and industry vs.\ government regulation); (4) \textbf{source selection} (e.g., ``leading scientists'' vs.\ ``alarmists''); and (5) \textbf{language and tone} (e.g., ``climate crisis'' vs.\ ``climate agenda''). Fall back on holistic, intuitive judgment only when these conflict or are absent, and note this in a comment.

\subsection*{Handling Ambiguous, Mixed, and Insufficient-Information Cases}

\begin{itemize}
    \item \textbf{Genuinely balanced.} The text presents left- and right-coded framings or sources without favoring either. Assign \textbf{0}; a comment is optional.
    \item \textbf{Insufficient or non-substantive content.} The scraped text is not a substantive article (e.g., navigation menu, author bio, copyright/legal boilerplate, an unrelated piece that only inherited the climate-change topic tag).
\end{itemize}

\subsection*{Annotation Procedure}

\begin{itemize}
    \item Do \textbf{not} conduct external research or fact-checking; base your score only on the text provided.
    \item Apply the indicators and decision rules above consistently across texts, rather than case-by-case intuition.
\end{itemize}

The aim is to capture the \textit{perceived ideological orientation} of each text as it might be interpreted in everyday political discourse, using the indicators above.

\subsection*{Coding Process and Intercoder Reliability}

Two independent annotators applied the codebook above. Both annotators rated all 120 articles in the sample. Coders were shown only the article text reproduced in the interface below; the outlet name, URL, and any other publication metadata were not displayed, so that ratings would reflect the text itself rather than prior beliefs about a given outlet's ideological reputation.

Figure~\ref{fig:coding_example} reproduces, for one article in the sample, exactly what a coder saw: the article text and the scoring prompt from the codebook above, with no outlet or author information attached.

\begin{figure}[h]
\begin{center}
\fbox{\begin{minipage}{0.85\linewidth}
\vspace{0.3cm}
\textbf{Article text:}

\textit{``Chasing Coral. As climate change deniers take power in the U.S., Jeff Orlowski's documentary Chasing Coral provides a needed corrective. In this companion film to Orlowski's Chasing Ice (2012), he and a team of scientists attempt to record the devastating effects of rising sea temperatures on coral, the foundational organism of the world's oceans. Former advertising executive Richard Vevers joined Orlowski and Zack Rago to create and deploy a waterproof imaging system capturing the declined in coral reefs in real time. The film promises spectacular photography documenting an environmental crisis. Screening April 6 at 6 p.m. at the Castro Theatre.''}

\vspace{0.3cm}
\textbf{Assign an ideology score (--5 far left to +5 far right):} \rule{2cm}{0.4pt}

\textbf{Comment (optional):} \rule{4cm}{0.4pt}
\vspace{0.2cm}
\end{minipage}}
\end{center}
\caption{Illustrative reconstruction of the coding interface for one article in the sample, based on the codebook in this appendix. Outlet, author, and URL were not shown to coders.}
\label{fig:coding_example}
\end{figure}

We assess intercoder reliability by comparing the two annotators' scores across all 120 articles. Agreement is substantial: Pearson $r = 0.81$; quadratic-weighted Cohen's $\kappa = 0.80$. The weighted measure accounts for the size of a disagreement rather than requiring an exact match, which is the appropriate criterion for an 11-point ideology scale. 54.2\% of articles receive identical scores from both coders, and 86.7\% are within one point of each other.

\subsection*{Composition of the Climate Change Article Pool}

\citet{Thomas2019Predicting} construct their corpus using a webly supervised approach: topic and ideology labels are assigned at the level of the source webpage based on its text, and every image appearing on a labeled page inherits that page's topic and ideology label without independent, per-image verification. As a result, a single webpage can contribute several images to the pool, not all of which depict the labeled topic directly; some labeled pages are author biography or archive listings rather than standalone articles. This is important context for interpreting the example images discussed below, several of which (e.g., film stills, product photography) are better understood as other images scraped from a page classified as climate change than as depictions of climate change itself.

Before drawing our stratified sample of 120 articles, 2,853 climate change images were available across the eight outlets. Table~\ref{table:cc_pool} reports how this pool is distributed across outlets. Because our sample fixes 15 images per outlet, the annotated ground truth is stratified by outlet rather than proportional to this underlying distribution.

\begin{table}[h]
\begin{center}
\begin{tabular}{l c c}
\hline
Outlet & Articles & Share \\
\hline
Newsmax   & 781 & 27.4\% \\
RedState  & 354 & 12.4\% \\
Advocate  & 349 & 12.2\% \\
Breitbart & 329 & 11.5\% \\
Wonkette  & 291 & 10.2\% \\
Fox News  & 269 &  9.4\% \\
CNN       & 247 &  8.7\% \\
Esquire   & 233 &  8.2\% \\
\hline
Total     & 2{,}853 & 100\% \\
\hline
\end{tabular}
\caption{Distribution of the climate change article pool across the eight outlets, prior to stratified sampling for hand coding.}
\label{table:cc_pool}
\end{center}
\end{table}

The largest category is portrait or headshot photography of politicians, pundits, and other public figures quoted or discussed in the article (e.g., a head of state, a television commentator, an activist or celebrity), which is typical of opinion journalism, where the image illustrates who is speaking rather than the subject matter itself. Other substantial shares depict protest and activism scenes (e.g., demonstrators holding signs, an advocacy group's staged photo opportunity), political-conference settings (e.g., a speaker at a podium), and genuine climate- or weather-related imagery (e.g., storm clouds, a stock illustration contrasting green land with cracked, drought-stricken ground). This heterogeneity is what makes climate change a useful test case for our framework: identifying which object in a given frame is most prominent, a face, a protest sign, a flag, or a landscape, is a genuinely non-trivial task across this sample.

\subsection*{External Validation Against Independent Media-Bias Rankings}

Our ideology scores rely on the average of two independent research assistants' judgments, following the codebook above (Appendix~\ref{sec:codebook}). To address the concern that this annotation might not track how these outlets are actually perceived, we validate the outlet-level averages of our hand-coded ground truth against two independent, publicly available media-bias rating services that do not draw on our annotation: Media Bias/Fact Check (MBFC) and Ad Fontes Media (AFM).\footnote{MBFC: \url{https://mediabiasfactcheck.com/}; AFM: \url{https://adfontesmedia.com/}. Both rate outlets on a left--right numeric scale using independent panels of analysts; scores retrieved July 2026. MBFC reports no numeric score for Esquire (a categorical ``Left'' rating only), so Esquire is excluded from the MBFC comparison ($n=7$) but included in the Ad Fontes comparison ($n=8$).}

Table~\ref{table:exogenous_validation} reports outlet-level averages alongside both external rankings. The hand-coded ground truth correlates strongly with each (MBFC: Pearson $r=0.92$, Spearman $\rho=0.96$; Ad Fontes Media: Pearson $r=0.95$, Spearman $\rho=0.90$). As a further check, we repeat this validation using only the original single annotator instead of the two-annotator average; the correlations remain equally strong (MBFC: Pearson $r=0.92$, Spearman $\rho=0.89$; Ad Fontes Media: Pearson $r=0.93$, Spearman $\rho=0.88$), indicating that the agreement with external rankings is not an artifact of averaging across annotators.

\begin{table}[h]
\begin{center}
\begin{tabular}{l c c c}
\hline
Outlet & Hand-Coded (Mean, SE) & MBFC & Ad Fontes Media \\
\hline
Wonkette   & $-2.70$ (0.20) & $-8.0$ & $-25.02$ \\
Advocate   & $-1.80$ (0.24) & $-6.4$ & $-12.43$ \\
Esquire    & $-1.13$ (0.38) & --     & $-18.65$ \\
CNN        & $-0.67$ (0.25) & $-3.6$ & $-6.27$  \\
Newsmax    & $0.13$ (0.35)  & $7.8$  & $13.32$  \\
Breitbart  & $0.60$ (0.42)  & $8.1$  & $13.57$  \\
Fox News   & $0.63$ (0.33)  & $8.0$  & $11.24$  \\
RedState   & $1.87$ (0.41)  & $8.5$  & $21.33$  \\
\hline
\end{tabular}
\caption{Outlet-level hand-coded ideology scores compared with two independent media-bias rankings. Hand-coded values are means (with standard errors, in parentheses) of our $-5$ (far left) to $+5$ (far right) annotation scale, computed over the 15 sampled articles per outlet; MBFC and Ad Fontes Media scores are each service's own numeric bias scale, not directly comparable in magnitude across services or with our scale.}
\label{table:exogenous_validation}
\end{center}
\end{table}

\subsection*{Text as Ground Truth: Possible Divergence from Image Content}

We adopt text as our ground truth by design, following the convention of the underlying corpus \citep{Thomas2019Predicting}, rather than by oversight. We recognize that text and image content within the same article need not always align in ideological valence or framing \citep{messaris2001role, grabebucy2009image}. To the extent that such divergence is present in our sample, it constitutes measurement error in the ground truth rather than in the object-prominence measure itself, which biases the estimated benefit of salience weighting toward zero rather than away from it. The improvement we report from adding prominence weights is therefore, if anything, a conservative estimate of the true relationship between object prominence and ideological positioning.

\section{Application 1: Matrix Construction, Hyperparameter Selection, and Robustness Checks}\label{sec:app1-hyperparams}

\subsection*{Worked Example: Object-Document Matrix Construction}

Consider, for illustration, a hypothetical image containing three regions that a detector might identify: a person occupying the center of the frame, a building in the background, and a stretch of sky. Under the supervised approach, Detectron2 assigns each region to a named category; under the raw, unweighted representation, each region simply contributes a count of one to its column of the object-document matrix, so all three appear equally important. Under the salience-weighted representation, each region instead contributes its own average MBD salience score, computed over that region's own pixels: a centered, high-contrast foreground figure might score highly (e.g., 150 on MBD's 0--255 scale), while the receding building and sky score much lower (e.g., 40 and 15)\textemdash reflecting that the person dominates the composition while the background recedes.

The unsupervised approach represents the same compositional information differently. Rather than named categories, RootSIFT locates a large number of local keypoints scattered across the image and assigns each to one of 2,000 visual-word clusters, using a codebook built once, corpus-wide, by $k$-means clustering in which each keypoint's own salience score enters directly as a clustering weight (scikit-learn's \texttt{sample\_weight} argument), so cluster centers are pulled toward the most salient keypoints across the corpus. For our hypothetical image, suppose several keypoints on the person's face and clothing are assigned to cluster 482, while keypoints on the building's facade are assigned to a different cluster, 791. Each keypoint's own salience is read directly from the pixel it sits on; the image's contribution to cluster 482 is the average salience of only that image's own keypoints assigned there (illustratively, high), while its contribution to cluster 791 is the average salience of the keypoints assigned there (illustratively, low). Because a single vocabulary of 2,000 clusters is shared across the whole corpus, cluster indices do not carry an inherent semantic label the way Detectron2's object categories do\textemdash cluster 482 is not intrinsically "person"\textemdash but weighting still differentiates a frequently salient region of an image from a frequently backgrounded one.

In both cases, these per-image, per-object values are summed within each outlet across all of that outlet's climate-change images (drawn from the full pool of images per outlet, not only the 120 hand-coded articles, which serve solely as the ground truth against which Table~\ref{tab:scaling_mae}'s MAE is computed) to form the rows of the outlet-by-object matrix that Wordfish scales. Wordfish then estimates, for outlet $i$ and object $j$ (a Detectron2 category or a visual-word cluster, depending on the approach), a model of the form $\log \lambda_{ij} = \alpha_i + \psi_j + \beta_j \theta_i$, where $\theta_i$ is the outlet's position on the latent left-right dimension reported in the main text, $\psi_j$ captures how frequently object $j$ appears overall, and $\beta_j$ captures how strongly object $j$ discriminates between left- and right-leaning outlets\textemdash the object-level analogue of a word's discriminating power in text scaling.

\subsection*{Hyperparameter Grid}

We try different vocabulary sizes for the model and exclude objects that are very rare or very frequent \citep{denny2018text, Arnold_Biedebach_Küpfer_Neunhoeffer_2024}. We run Wordfish separately for each combination of hyperparameters using the full dataset. For the main results (Table 1), we report the average estimates across all runs. 

\begin{itemize}
    \item \texttt{min\_freq}: [0.0000, 0.0001, 0.0002, 0.0003, 0.0004, 0.0005]
    \item \texttt{max\_freq}: [0.05, 0.10, 0.15, 0.20, 0.25, 0.30, 0.35, 0.40, 0.45, 0.50, 1.00]
\end{itemize}

These are the best settings per scenario (\texttt{min\_freq}, \texttt{max\_freq}):
\begin{itemize}
    \item Unsupervised: 0.0000, 0.05
    \item Unsupervised (Prominence): 0.0004, 0.05
    \item Supervised: 0.0005, 0.45
    \item Supervised (Prominence): 0.0002, 0.05
\end{itemize}

\section{Application 2: Data Processing and Regression Results}\label{sec:app2-data}

\subsection*{Data Processing and Sample Composition}

We process 1,917 video files (823 from 2016, 1,094 from 2020), applying scene-change detection to extract 40,370 scene frames (Table~\ref{table:processing_funnel}). For context, the Wesleyan Media Project (WMP) video-level metadata holds information on all presidential campaign ads across the two cycles. We then run face detection and gender classification on all extracted frames, yielding 98,318 face-level observations in total.

For the regression models reported below, we restrict this raw set of observations to ads sponsored by a Democratic or Republican candidate, frames with a genuinely detected face (as opposed to the full-frame bounding box that the face-detection model returns when no face is found), and faces classified as male or female with at least 60\% confidence. This final analytic sample comprises 1,721 videos and 22,620 unique frames. Of the resulting 64,568 face-level observations, 162 additionally lack the candidate-visibility indicator used as a fixed effect and are dropped during estimation, leaving 64,406 observations for the Depth Model; the Relative Face Size Model further requires the on-screen candidate's gender to be known, leaving 58,674 observations.

\begin{table}[h]
\begin{center}
\begin{tabular}{l c c c}
\hline
 & WMP Videos Processed & Scene Frames Extracted \\
\hline
2016  & 823   & 12,926   \\
2020  & 1,094 & 27,444 \\
\hline
Total & 1,917 & 40,370 \\
\hline
\end{tabular}
\caption{Video-to-image processing funnel by election cycle. The rightmost column is shown for context only; it does not subtract exactly from Videos Processed (see main text).}
\label{table:processing_funnel}
\end{center}
\end{table}

Table~\ref{table:party_balance} reports the resulting balance between Democratic and Republican ads in the final analytic sample. The 2016 subsample is close to evenly split between the two parties, whereas the 2020 subsample skews Democratic (79\% of videos), reflecting the large field of Democratic primary candidates competing for airtime that cycle relative to a largely unopposed Republican incumbent.

\begin{table}[h]
\begin{center}
\begin{tabular}{l c c c c c c}
\hline
 & \multicolumn{2}{c}{Videos} & \multicolumn{2}{c}{Frames} & \multicolumn{2}{c}{Face Observations} \\
Election & Dem. & Rep. & Dem. & Rep. & Dem. & Rep. \\
\hline
2016  & 386   & 388 & 4,899  & 3,510 & 19,865 & 8,403 \\
2020  & 751   & 196 & 11,597 & 2,614 & 30,361 & 5,939 \\
\hline
Total & 1,137 & 584 & 16,496 & 6,124 & 50,226 & 14,342 \\
\hline
\end{tabular}
\caption{Balance between Democratic and Republican ads, by election cycle, prior to dropping 162 observations lacking the candidate-visibility indicator used in the Depth Model (leaving a final N of 64,406; see main text).}
\label{table:party_balance}
\end{center}
\end{table}

\begin{figure}[h]
    \centering
    \includegraphics[width=.8\linewidth]{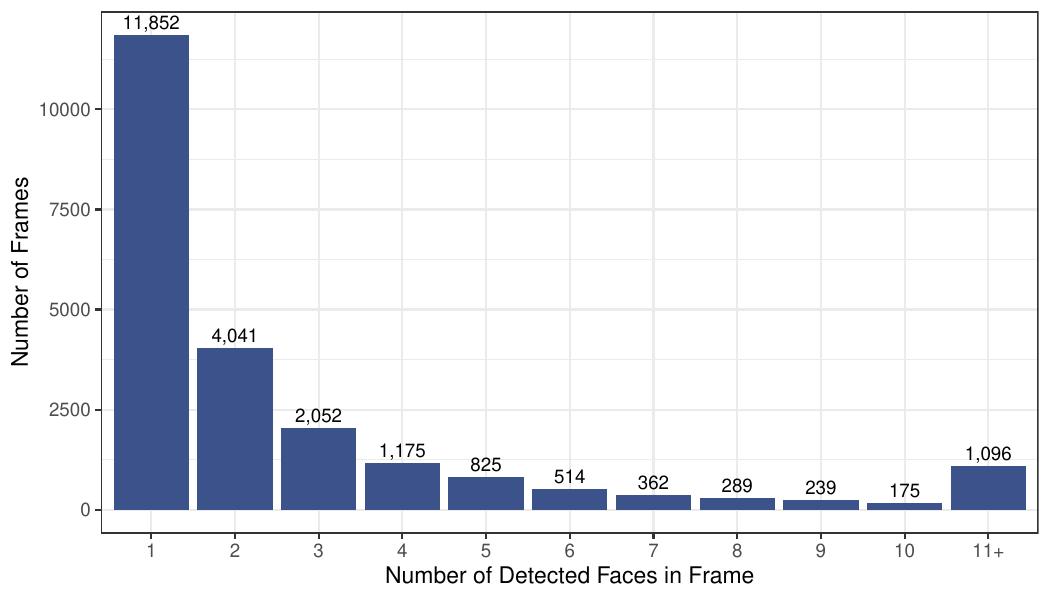}
    \caption{Distribution of the number of detected faces per frame in the final analytic sample.}
    \label{fig:faces_per_frame}
\end{figure}

Figure~\ref{fig:faces_per_frame} reports the distribution of the number of detected faces per frame in the final analytic sample. Only 52.4\% of the 22,620 frames contain a single face; the remainder contain two or more, and about 5\% contain 11 or more faces, reflecting rally and crowd scenes. Salience weighting is only trivial when a frame contains exactly one candidate object; since roughly half of our frames contain multiple faces, choosing which face is most prominent is a substantive task in the majority of cases where it applies.

\begin{figure}[h]
    \centering
    \includegraphics[width=.8\linewidth]{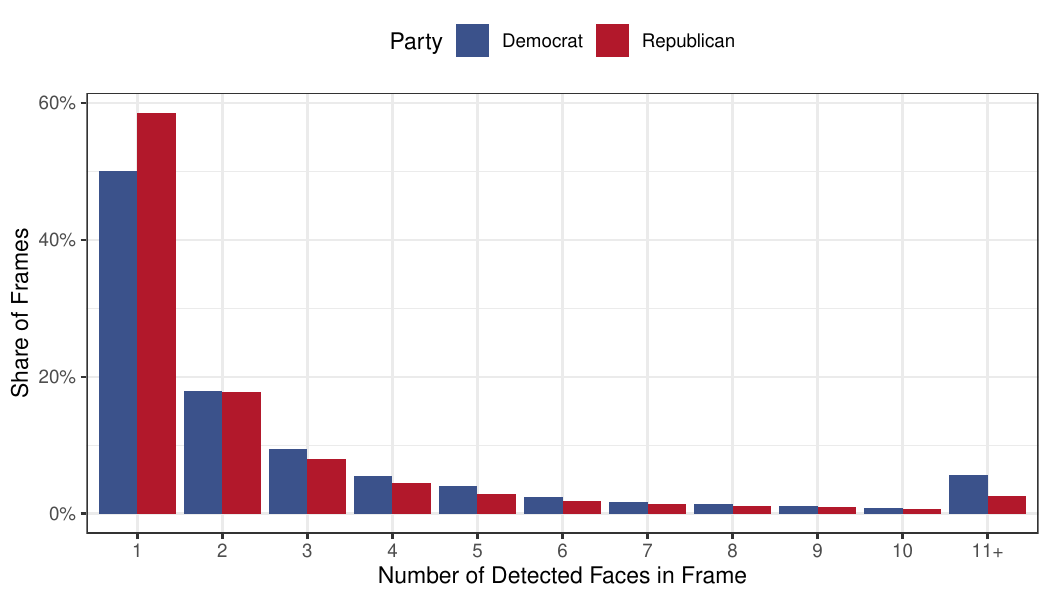}
    \caption{Distribution of the number of detected faces per frame in the final analytic sample, by party.}
    \label{fig:faces_per_frame_party}
\end{figure}

For completeness, Figure~\ref{fig:faces_per_frame_party} reports the same distribution separately for Democratic and Republican frames.

\subsection*{Regression Results}

\begin{table}[h]
\begin{center}
\begin{tabular}{l c c}
\hline
 & Depth Model & Face Size Model \\
\hline
Intercept                          & $-0.12$     & $0.49$      \\
                                   & $(0.02)$    & $(0.06)$    \\
Gender: Female                     & $-0.01$     & $0.14$      \\
                                   & $(0.01)$    & $(0.01)$    \\
Party: Republican                  & $-0.27$     & $-0.37$     \\
                                   & $(0.02)$    & $(0.09)$    \\
Gender: Female × Party: Republican & $-0.10$     & $-0.19$     \\
                                   & $(0.02)$    & $(0.03)$    \\
\hline
Num. obs.                          & 64406       & 58674       \\
Fixed Effects: Candidate           & \checkmark           & \checkmark           \\
Fixed Effects: Candidate Visible   & \checkmark           & \checkmark           \\
Fixed Effects: Election            & \checkmark          & \checkmark           \\
AIC                                & $165757.98$ & $155281.38$ \\
BIC                                & $166284.21$ & $155775.27$ \\
\hline
\end{tabular}
\caption{Statistical models for estimating prominence in presidential video ad frames.}
\label{table:coefficients}
\end{center}
\end{table}

\subsection*{Robustness: Number of Faces per Frame}

Table~\ref{table:robustness_faces} refits the Gender $\times$ Party interaction restricting the sample to frames with two or more detected faces, to assess whether the reported effects are driven by frames in which prominence trivially reduces to a single candidate object. For the Depth Model, the interaction is essentially unchanged ($-0.12$, compared to $-0.10$ in the full sample). For the Face Size Model, the interaction is roughly halved ($-0.09$, compared to $-0.19$ in the full sample) but remains negative and statistically significant. In both models, the qualitative conclusion is unaffected by this restriction.

\begin{table}[h]
\begin{center}
\begin{tabular}{l l c c c}
\hline
Model & Subsample & Estimate & SE & N \\
\hline
Depth Model      & Full sample   & $-0.10$ & $0.02$ & 64,406 \\
Depth Model      & 2+ faces only & $-0.12$ & $0.03$ & 52,618 \\
Face Size Model  & Full sample   & $-0.19$ & $0.03$ & 58,674 \\
Face Size Model  & 2+ faces only & $-0.09$ & $0.02$ & 49,044 \\
\hline
\end{tabular}
\caption{Gender $\times$ Party interaction (Female $\times$ Republican) by number of detected faces per frame.}
\label{table:robustness_faces}
\end{center}
\end{table}

\begin{figure}[h]
    \centering
    \includegraphics[width=.8\linewidth]{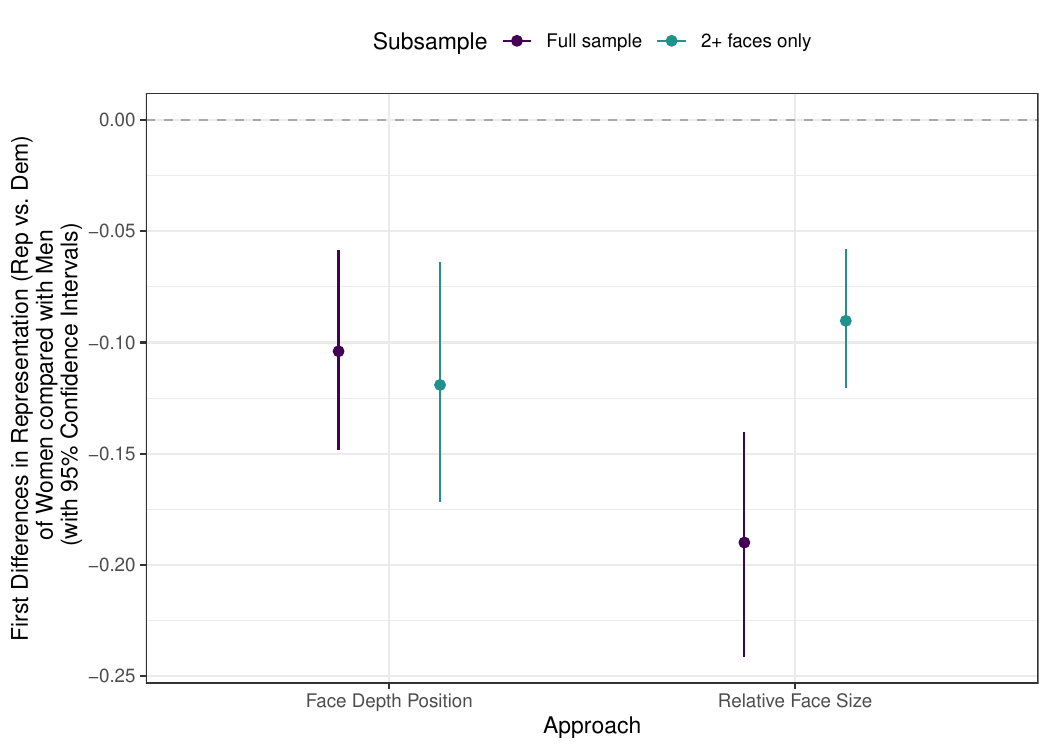}
    \caption{First Differences of Expected Values (as in Figure~\ref{fig:results}), estimated separately for the full sample and for frames with 2+ detected faces.}
    \label{fig:results_by_faces}
\end{figure}

Figure~\ref{fig:results_by_faces} presents the same comparison in the form of Figure~\ref{fig:results} in the main text, regenerating the simulation-based first differences for both model specifications (Face Depth Position, Relative Face Size) separately for the full sample and for frames with 2+ detected faces.

\newpage
\section{Validation: First Fixation, Algorithm Sensitivity, and Deviations}\label{app:validation}

This appendix collects the material relegated from Section~\ref{sec:validation}: the second experiment on ideological moderation, the first-fixation measure, the sensitivity of the alignment to the choice of operationalization, the deviations from the pre-analysis plan, and the full coefficient tables. All models are estimated at the cell level (participant $\times$ image $\times$ cell) on the pre-registered $4\times4$ grid; the main text additionally reports the finer $8\times8$ grid.

\subsection*{Experiment 2: do left- and right-leaning viewers attend differently?}
The pre-analysis plan framed Experiment~2 around an interaction between the image's outlet and the viewer's ideology, on the expectation that outlets of opposing political leanings would engage viewers of opposing ideologies differently. That expectation does not survive contact with the stimuli: both outlets we sampled, the Guardian and the Daily Mirror, sit on the political left. Do left- and right-leaning \emph{viewers} differ in how they attend to this imagery? Fifty-three of the same participants viewed seventeen news photographs of three highly politicized British events\textemdash the Southport protests of July 2024, the right-wing protests in London of September 2025, and reporting on small-boat migration in February 2026\textemdash presented in matched Guardian--Mirror pairs so that date and event is held fixed. Because both outlets lean left and the events are charged (protest and migration), the corpus is, if anything, slanted. This makes the comparison a conservative one, since it is exactly where a viewer's politics would be most likely to shape where they look.

We split the $47$ participants who reported their ideology at the sample median of a self-reported $0$--$10$ left--right scale (median $=3$), giving $27$ left-of-median and $20$ right-of-median viewers, with ties assigned to the left. For each participant and each dependent variable we form a \emph{spatial profile}: the participant's mean value (fixation count, dwell time, or returns) in each cell of the imposed grid, pooled across the images they saw. Two questions must be kept distinct. First, whether one group attends \emph{more overall} (the level of attention). Second, whether the groups distribute attention \emph{differently across the grid} (the spatial pattern) irrespective of the total. We test both. Each test compares the two groups over the $47$ participants rather than over the individual cells, so that the many correlated cells within a viewing are not treated as independent observations.

The first test is a two-sample $t$-test on each participant's overall mean per cell. It answers only the level question: does one group look more in total? The second is \emph{Hotelling's $T^2$}, the multivariate generalization of that $t$-test. Rather than compare a single number, it compares the two groups' entire mean profiles across the grid at once, while accounting for the correlations among cells. We report it in two forms. The \emph{raw} version is sensitive to both the level and the spatial pattern. The \emph{shape} version first centres each participant's profile on its own mean, so that only the pattern\textemdash where attention goes, independent of how much\textemdash enters the comparison. Hotelling's $T^2$ requires more participants than cells (the pooled covariance must be invertible), so it is available on the $4\times4$ grid ($16$ cells, $47$ participants) but not on the $8\times8$ grid ($64$ cells), where the covariance is singular.

The third test is a \emph{permutation test}, which removes the dimensionality limit and makes no distributional assumption\textemdash useful here, since the measures are bounded counts and times on a fixed budget. We summarize the group difference by a single statistic, the summed squared difference, cell by cell, between the two groups' mean profiles, and we ask how extreme the observed value is under the null that a viewer's ideology is unrelated to the profile. Under that null the labels \emph{left} and \emph{right} are exchangeable across participants, so we recompute the statistic for $5{,}000$ random re-assignments of the labels, each time moving each participant's entire profile as one block, and record the fraction of these that match or exceed the observed value.

Table~\ref{tab:val_h2} collects the results. On no measure, grid, or test is there a significant difference between left- and right-leaning viewers. The level $t$-tests are far from significance ($p\ge0.17$). The permutation tests are non-significant everywhere (all $p\ge0.13$). Hotelling's $T^2$ on the $4\times4$ grid agrees; its one near-threshold value, the raw fixation-count profile ($F(16,30)=1.84$, $p=0.073$), combines level and pattern and is contradicted by both the shape-only version ($p=0.13$) and the corresponding permutation test ($p=0.16$), so we read it as noise rather than signal. For completeness, the pre-registered outlet~$\times$~ideology interaction is likewise null on every measure and grid.

We draw a cautious conclusion. Within this sample, viewers' politics do not detectably change either how much or where they look at charged political imagery, which is consistent with the framework's claim that the composition of an image, rather than the viewer's ideology, governs the allocation of attention. This is an absence of evidence rather than proof of invariance: arguing positively for no effect would require a properly powered equivalence test, and our power is limited\textemdash there are $20$ right-of-median viewers, and Hotelling's $T^2$ estimates fifteen or sixteen profile dimensions from $47$ people. The sample also leans left (median $3$, about two-thirds left of centre), so the comparison speaks to the ideological range present rather than to strongly right-wing viewers. Within those limits, we find no sign that prominence lies in the eye of the partisan beholder.

\begin{table}[!htbp]
\centering
\caption{Experiment 2: left- versus right-leaning viewers across the grid. Entries are $p$-values for a difference between the two viewer groups; larger is more consistent with no difference.}
\label{tab:val_h2}
\begin{threeparttable}
\begin{tabular}{lccccc}
\toprule
 & \multicolumn{1}{c}{Level} & \multicolumn{2}{c}{Hotelling's $T^2$} & \multicolumn{2}{c}{Permutation} \\
\cmidrule(lr){2-2}\cmidrule(lr){3-4}\cmidrule(lr){5-6}
 & \multicolumn{1}{c}{$t$-test} & \multicolumn{1}{c}{raw} & \multicolumn{1}{c}{shape} & \multicolumn{1}{c}{raw} & \multicolumn{1}{c}{shape} \\
\midrule
\multicolumn{6}{l}{\emph{Panel A: registered $4\times4$ grid}} \\[2pt]
Fixation count & 0.174 & 0.073 & 0.131 & 0.164 & 0.183 \\
Dwell time     & 0.311 & 0.100 & 0.133 & 0.191 & 0.191 \\
Returns        & 0.858 & 0.326 & 0.329 & 0.183 & 0.128 \\
\midrule
\multicolumn{6}{l}{\emph{Panel B: $8\times8$ grid}} \\[2pt]
Fixation count & 0.174 & --- & --- & 0.273 & 0.298 \\
Dwell time     & 0.311 & --- & --- & 0.329 & 0.327 \\
Returns        & 0.481 & --- & --- & 0.221 & 0.192 \\
\bottomrule
\end{tabular}
\begin{tablenotes}[flushleft]\footnotesize
\item Comparison of left-of-median ($n=27$) and right-of-median ($n=20$) viewers, split at the sample median of self-reported ideology ($=3$ on a $0$--$10$ left--right scale; ties assigned to the left). Each participant's spatial profile is the mean of the dependent variable per grid cell, pooled across the images they viewed. \emph{Level}: two-sample $t$-test on the participant's overall mean per cell (does one group attend more in total?). \emph{Hotelling's $T^2$}: multivariate two-sample test comparing the full per-cell profiles; the \emph{raw} version compares level and spatial pattern jointly, the \emph{shape} version compares the spatial pattern only (each profile first centred on its own mean). Hotelling's $T^2$ is undefined on the $8\times8$ grid, where the number of cells ($64$) exceeds the residual degrees of freedom ($45$), so the pooled covariance is singular. \emph{Permutation}: $5{,}000$ label permutations of the summed squared cell-wise difference between the two group-mean profiles (permuting the two viewer groups), for raw and shape profiles. No test reaches significance.
\end{tablenotes}
\end{threeparttable}
\end{table}

\subsection*{First fixation}
The pre-analysis plan registered a fourth attention measure, the cell that received the viewer's very first fixation. Unlike the three count measures, it shows no reliable alignment with algorithmic prominence: the interaction coefficient is positive but far from significant (column~4 of Table~\ref{tab:val_h1_itti}). We do not read this as evidence against the framework. The first fixation is a single, noisy event\textemdash early in viewing the eye tends to wander before it settles\textemdash so that a systematic effect on one initial landing point is far harder to detect than an effect on measures that accumulate over the whole viewing period. Establishing it would require a considerably larger sample.

\subsection*{Sensitivity to the choice of operationalization}
\begin{figure}[htbp]
    \centering
    \includegraphics[width=\linewidth]{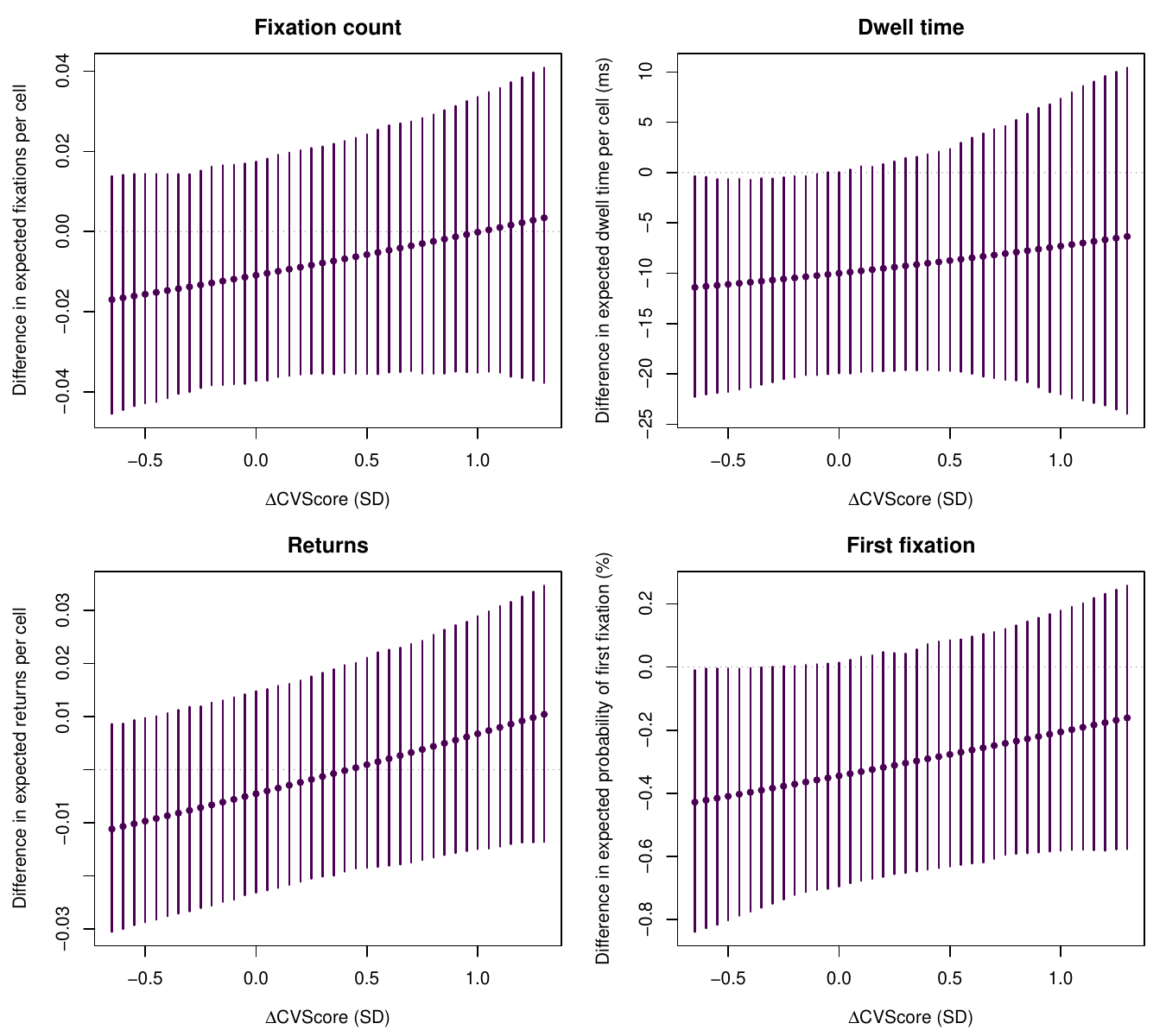}
    \caption{Experiment 1, re-estimated with MBD prominence. Simulated first differences in expected attention across the range of $\Delta\text{CVScore}$. Unlike the Itti model (Figure~\ref{fig:val_h1}), MBD produces no reliable alignment with human attention on any measure.}
    \label{fig:val_mbd}
\end{figure}

The alignment in Experiment~1 depends on how prominence is measured. As explained in Section~\ref{sec:validation}, salient object detection comes in two strands: algorithms built to mimic human gaze, such as the Itti model used above, and algorithms built to segment the compositionally most salient object, such as minimum barrier detection (MBD). Only the first strand is designed to predict where people look. Re-running Experiment~1 with MBD confirms the distinction: none of the interaction coefficients is statistically distinguishable from zero, and the first-difference curves stay flat across the range of predicted prominence (Figure~\ref{fig:val_mbd}; Table~\ref{tab:val_h1_mbd}). This is exactly what should be expected, and it is not a defect of MBD. MBD is a segmentation method, tuned to isolate a coherent foreground object rather than to reproduce fixations; that it does not track gaze is by design, not a failure of the framework. It also means the result does not undercut the applications, where MBD serves as a measure of compositional salience and makes no claim to model gaze. The lesson for practice is simply that, when the goal is alignment with human attention, a gaze-mimicking saliency model is the operationalization to use\textemdash which is the one we validate here. Size, centeredness, and depth were not part of this test and remain to be assessed against gaze.

\subsection*{Deviations from the pre-analysis plan}
We registered the design before collecting data and report all departures from it. The pre-registered analysis used a $4\times4$ grid; we additionally report a finer $8\times8$ grid, which improves spatial resolution and, if anything, strengthens the Experiment~1 estimates. Experiment~2 used seventeen images rather than the registered twelve, with image fixed effects absorbing the difference. Dwell time is modeled in units of 100\,ms rather than raw milliseconds, because the raw counts are too dispersed for the negative binomial. The first-fixation model uses a logit with a leave-one-out attention control in place of the registered conditional logit, because cells that are never first-fixated cause separation under cell fixed effects. First fixation is defined as the first fixation beginning after image onset, so the residual central fixation left by drift correction is excluded. In the Experiment~2 count models we replace image fixed effects with an outlet indicator so that the outlet effect is estimable. Two substantive qualifications also follow from the analysis: the alignment in Experiment~1 is not uniform across the seven images, and, as noted above, it depends on the choice of algorithm.

\subsection*{Coefficient tables}
\begin{table}[!htbp]
\centering
\caption{Experiment 1: Human attention and algorithmic prominence, Itti model (registered $4\times4$ grid)}
\label{tab:val_h1_itti}
\begin{threeparttable}
\begin{tabular}{lcccc}
\toprule
 & \multicolumn{1}{c}{Fixation count} & \multicolumn{1}{c}{Dwell time} & \multicolumn{1}{c}{Returns} & \multicolumn{1}{c}{First fixation} \\
 & (1) & (2) & (3) & (4) \\
\midrule
Treatment & 0.002 & 0.008 & 0.006 & -0.144 \\
 & (0.032) & (0.040) & (0.026) & (0.127) \\[2pt]
$\Delta$CVScore (std.) & -0.228$^{***}$ & -0.276$^{***}$ & -0.180$^{***}$ & 0.170 \\
 & (0.023) & (0.028) & (0.019) & (0.087) \\[2pt]
Treatment $\times$ $\Delta$CVScore & \textbf{0.092$^{**}$} & \textbf{0.133$^{***}$} & \textbf{0.101$^{***}$} & \textbf{0.170} \\
 & (0.031) & (0.038) & (0.026) & (0.108) \\[2pt]
\midrule
Participant FE & Yes & Yes & Yes & Yes \\
Image FE & Yes & Yes & Yes & Yes \\
Cell FE & Yes & Yes & Yes & \multicolumn{1}{c}{LOO} \\
Model & Neg. bin. & Neg. bin. & Neg. bin. & Logit \\
Observations & 6,016 & 6,016 & 6,016 & 6,016 \\
\bottomrule
\end{tabular}
\begin{tablenotes}[flushleft]\footnotesize
\item Cell-level observations (participant $\times$ image $\times$ cell). $\Delta$CVScore standardized. The interaction coefficient tests H1. First fixation: binomial logit; cell fixed effects are replaced by a leave-one-out attention control (mean dwell of all other participants on the cell, standardized), since cells never first-fixated cause separation. Dwell time is modeled in units of 100\,ms (approximately the minimum fixation duration). Standard errors in parentheses. $^{*}p<0.05$; $^{**}p<0.01$; $^{***}p<0.001$.
\end{tablenotes}
\end{threeparttable}
\end{table}

\begin{table}[!htbp]
\centering
\caption{Experiment 1 with MBD prominence: no alignment with human attention (registered $4\times4$ grid)}
\label{tab:val_h1_mbd}
\begin{threeparttable}
\begin{tabular}{lcccc}
\toprule
 & \multicolumn{1}{c}{Fixation count} & \multicolumn{1}{c}{Dwell time} & \multicolumn{1}{c}{Returns} & \multicolumn{1}{c}{First fixation} \\
 & (1) & (2) & (3) & (4) \\
\midrule
Treatment & -0.008 & -0.018 & 0.005 & -0.083 \\
 & (0.032) & (0.041) & (0.026) & (0.125) \\[2pt]
$\Delta$CVScore (std.) & 0.025 & 0.032 & 0.007 & 0.016 \\
 & (0.018) & (0.026) & (0.013) & (0.056) \\[2pt]
Treatment $\times$ $\Delta$CVScore & \textbf{-0.021} & \textbf{-0.023} & \textbf{0.010} & \textbf{0.035} \\
 & (0.026) & (0.036) & (0.018) & (0.079) \\[2pt]
\midrule
Participant FE & Yes & Yes & Yes & Yes \\
Image FE & Yes & Yes & Yes & Yes \\
Cell FE & Yes & Yes & Yes & \multicolumn{1}{c}{LOO} \\
Model & Neg. bin. & Neg. bin. & Neg. bin. & Logit \\
Observations & 6,016 & 6,016 & 6,016 & 6,016 \\
\bottomrule
\end{tabular}
\begin{tablenotes}[flushleft]\footnotesize
\item Cell-level observations (participant $\times$ image $\times$ cell). Specification identical to Table~\ref{tab:val_h1_itti}, but prominence is measured with minimum barrier detection (MBD) rather than the Itti model. No interaction coefficient is statistically distinguishable from zero. Standard errors in parentheses. $^{*}p<0.05$; $^{**}p<0.01$; $^{***}p<0.001$.
\end{tablenotes}
\end{threeparttable}
\end{table}

\end{document}